\documentclass[letterpaper, 10 pt, conference]{ieeeconf}  % Comment this line out if you need a4paper

\IEEEoverridecommandlockouts                              % This command is only needed if 
\usepackage[acronyms]{glossaries}
\usepackage[utf8]{inputenc} % allow utf-8 input
\usepackage{hyperref}       % hyperlinks
\usepackage{url}            % simple URL typesetting
\usepackage{booktabs}       % professional-quality tables
\usepackage{amsfonts}       % blackboard math symbols
\usepackage{nicefrac}       % compact symbols for 1/2, etc.
\usepackage{microtype}      % microtypography
\usepackage{xcolor}         % colors
\usepackage{amsmath}
\usepackage{graphicx}
\usepackage{multirow}
\usepackage{tabularx}
\usepackage{graphicx}
\usepackage{colortbl}
\usepackage[table,xcdraw]{xcolor}
\usepackage{caption}
\usepackage{makecell}
\usepackage{cleveref}
\newacronym{3dgs}{3DGS}{3D Gaussian Splatting}
\newacronym{mlp}{MLP}{Multi-Layer Perceptron}
\newacronym{pe}{PE}{Positional Encoding}
\usepackage{float}

\title{\LARGE \bf
Neural-MMGS: Multi-modal Neural Gaussian Splats for Large-Scale Scene Reconstruction 
}

\author{Sitian Shen$^{1*}$, Georgi Pramatarov$^{1}$, Yifu Tao$^{2}$, Daniele De Martini$^{1}$% <-this % stops a space
\thanks{$^{1}$Sitian Shen, Georgi Pramatarov, and Daniele De Martini are with Mobile Robotics Group (MRG), Oxford Robotics Institute, Department of Engineering Science, University of Oxford, UK.
        {\tt\small \{sitian, daniele, georgi\}@robots.ox.ac.uk}}%
\thanks{$^{2}$Yifu Tao is with Dynamic Robot Systems Group (DRS), Oxford Robotics Institute, Department of Engineering Science, University of Oxford, UK.
        {\tt\small yifu@robots.ox.ac.uk}}%
}

\begin{document}

\maketitle
\thispagestyle{empty}
\pagestyle{empty}

%%%%%%%%%%%%%%%%%%%%%%%%%%%%%%%%%%%%%%%%%%%%%%%%%%%%%%%%%%%%%%%%%%%%%%%%%%%%%%%%
\begin{abstract}

This paper proposes Neural-MMGS, a novel neural \gls{3dgs} framework for multimodal large-scale scene reconstruction that fuses multiple sensing modalities in a per-gaussian compact, learnable embedding.
While recent works focusing on large-scale scene reconstruction have incorporated LiDAR data to provide more accurate geometric constraints, we argue that LiDAR's rich physical properties remain underexplored.
Similarly, semantic information has been used for object retrieval, but could provide valuable high-level context for scene reconstruction.
Traditional approaches append these properties to Gaussians as separate parameters, increasing memory usage and limiting information exchange across modalities.
Instead, our approach fuses all modalities -- image, LiDAR, and semantics -- into a compact, learnable embedding that implicitly encodes optical, physical, and semantic features in each Gaussian.
We then train lightweight neural decoders to map these embeddings to Gaussian parameters, enabling the reconstruction of each sensing modality with lower memory overhead and improved scalability.
We evaluate Neural-MMGS on the Oxford Spires and KITTI-360 datasets.
On Oxford Spires, we achieve higher-quality reconstructions, while on KITTI-360, our method reaches competitive results with less storage consumption compared with current approaches in LiDAR-based novel-view synthesis.

\end{abstract}

\glsresetall
%%%%%%%%%%%%%%%%%%%%%%%%%%%%%%%%%%%%%%%%%%%%%%%%%%%%%%%%%%%%%%%%%%%%%%%%%%%%%%%%
\section{INTRODUCTION}

Advancements in 3D scene reconstruction and view synthesis have been significantly impacted by the emergence of \gls{3dgs}~\cite{kerbl2023gaussiansplatting}.
This technique enables efficient, photorealistic rendering of large-scale environments by modelling scenes with spatially distributed Gaussian primitives with learnable attributes (e.g., colour, opacity and shape).
\Gls{3dgs} offers a highly compact and differentiable representation well-suited for training and deployment in constrained systems, as in robotics operations, where memory and latency are as critical as accuracy.

However, most existing \gls{3dgs} approaches rely primarily on RGB images, limiting their performance in challenging environments with poor lighting, occlusions, or textureless surfaces.
In contrast, robotic systems are inherently multimodal.
Cameras provide high-resolution texture and appearance, while range sensors such as LiDAR offer precise geometric structure.
More recently, pre-trained vision-language models offer semantic features -- e.g. CLIP embeddings~\cite{radford2021clip} -- that encode high-level, queryable concepts for downstream tasks.

Prior works \cite{qin2024langsplat, wu2024opengaussian, liao2024clip-gs} have begun incorporating LiDAR and semantic cues into \glspl{3dgs}.
However, these efforts often treat additional modalities as auxiliary, decoupled parameters, appended to each Gaussian.
This design choice not only increases memory usage but crucially misses the opportunity to leverage each sensor's complementary strengths.

We address these limitations with Neural-MMGS, a multimodal neural \gls{3dgs} framework that learns a unified, compact embedding per Gaussian, fusing information from RGB images, LiDAR point clouds, and semantic features.
This embedding captures optical, physical, and semantic cues simultaneously, enabling both cross-modal information exchange and efficient compression.
Instead of explicitly storing each modality's attributes, we train lightweight neural decoders that map the shared embedding to the parameters needed for rendering and reconstructing each domain.

This fused representation not only reduces memory overhead, but also facilitates synergistic learning: image-based appearance benefits from LiDAR’s geometric precision; geometry is guided by semantics (e.g., class-consistent priors); and semantic decoding is grounded in the actual geometry and appearance of the scene. Together, this design enables Neural-MMGS to produce richer, more consistent reconstructions across modalities.

To summarise, our main contributions are as follows:
\begin{enumerate}
    \item We introduce Neural-MMGS, a neural \gls{3dgs} framework that incorporates the optical, physical, and semantic features in a single per-Gaussian embedding;
    \item A modular architecture with modality-specific decoders, for joint reconstruction and synthesis of RGB, LiDAR and semantic information;
    \item A memory-efficient representation that simplifies the integration of diverse modalities without incurring high computational cost.
\end{enumerate}

We evaluate our method on Oxford Spires~\cite{tao2024oxford} and KITTI-360~\cite{liao2022kitti}, two large-scale, multimodal datasets with varying scene complexity and sensor characteristics.
Neural-MMGS performs well in both novel view synthesis and cross-modal rendering, achieving competitive visual quality, generalisation, and memory scalability.

\section{Related Work}

% This work intersects three key domains: Neural Gaussian Representation, Multi-Modal Supervision for \gls{3dgs}, and LiDAR Simulation.
% This section provides a brief overview of relevant works within each field, highlighting those that have particularly influenced our approach, and discusses how our proposed method differs from and builds upon existing research.

\subsection{Neural Gaussian Representation}
Although \gls{3dgs} achieves superior rendering quality and speed compared to NeRF-based methods, it typically requires hundreds of megabytes to store the attributes of 3D Gaussians.
This presents scalability challenges in large-scale scenes, especially when incorporating additional properties for each Gaussian point.
To address this, prior works \cite{ali2025elmgs,navaneet2024compgs,fan2024trim,fan2024lightgaussian,lee2024compact,niedermayr2024compressed,xie2024mesongs} have proposed various compression strategies, including pruning to reduce the number of Gaussians, vector quantisation to discretise attributes into shared codebooks, and context-aware entropy encoding.
More recent approaches \cite{tang2025neuralgs,pan2025pings, mihajlovic2024splatfields} aim to embed \gls{3dgs} into neural representations implicitly.
These methods attempt to bridge the gap between NeRF and \gls{3dgs} by leveraging the rendering efficiency of \gls{3dgs} and the compactness of neural storage. Beyond optical reconstruction, neural \gls{3dgs} has also been applied to LiDAR simulation.
For instance, LiDAR-GS \cite{chen2024lidargs}, built upon Scaffold-GS \cite{lu2024scaffold}, decodes learned embeddings into LiDAR-specific parameters tailored to point-level physical properties.

None of the above methods offers a unified embedding that jointly represents each Gaussian point's optical, semantic, and physical attributes.
SplatAD \cite{hess2024splatad} renders two separate embeddings into image and LiDAR view spaces, which are then decoded into corresponding parameter sets for RGB and LiDAR range maps.
Nonetheless, semantic information is not encoded, and the features remain modality-specific -- split between camera and LiDAR -- hindering effective cross-modal information fusion.

\subsection{Multi-Modal Supervision for \gls{3dgs}}
% \textcolor{red}{Add more cites here.}
Recent advances in \gls{3dgs} \cite{song2025adgaussian, cui2024letsgo, lee2025geomgs, kung2024lihi, shen2024draggaussian, wu2024mm} have explored multimodal supervision to enhance reconstruction quality.
ADGaussian \cite{song2025adgaussian} proposes a multi-modal learning framework that jointly optimises visual and sparse LiDAR features through a multi-scale Gaussian decoder.
LiDAR-based methods such as LetsGo~\cite{cui2024letsgo} and GeomGS ~\cite{lee2025geomgs} integrate accurate geometric signals for improved localisation and structural fidelity. In parallel, semantic supervision has been employed to enrich scene understanding.
LiHi-GS~\cite{kung2024lihi}, for instance, combines LiDAR, RGB, and semantic labels for autonomous driving scenes, but suffers from scalability issues due to explicit Gaussian parameterisation and underutilises LiDAR intensity information.

While prior work often treats each modality independently or appends them as separate attributes, our proposed Neural-MMGS takes a unified approach.
Our neural 3D Gaussian Splatting framework embeds multi-modal signals—image, LiDAR, and semantic data—into a compact, learnable per-Gaussian representation.

\subsection{LiDAR Simulation}

Recent data-driven approaches have advanced LiDAR simulation by reconstructing environments directly from real-world data, enabling high-fidelity and diverse scene representations that closely align with real-world distributions.
For example, LiDARsim \cite{manivasagam2020lidarsim} and PCGen \cite{li2023pcgen} simulate realistic point clouds through multi-step pipelines grounded in sensor recordings.
While effective, the complexity of these pipelines often hinders scalability and practical deployment.
To address these limitations, recent works \cite{tao2024lidarnerf,zhang2024nerflidar,zheng2024lidar4d,xue2024geonlf,tao2024alignmif,wu2024dynamiclidarsim} have explored the use of NeRF for LiDAR simulation, leveraging its neural radiance field representation to reconstruct photorealistic scenes.
However, NeRF-based methods are computationally expensive due to dense ray marching and are less suitable for large-scale, non-symmetric driving scenes.

To further improve efficiency and realism, recent studies such as GS-LiDAR~\cite{jiang2025gslidar}, LiDAR-RT~\cite{zhou2024lidarRT} and Industrial-GS~\cite{zeng2025industrial} adopt \gls{3dgs} for LiDAR simulation.
GS-LiDAR utilises panoramic Gaussian rendering combined with periodic vibration modeling~\cite{chen2023periodic}, producing temporally consistent and physically plausible LiDAR scans.
Building on this line of work, our method further integrates LiDAR simulation with multi-modal learning by embedding physical, semantic, and visual cues directly into compact per-Gaussian representations.
This allows us to simulate LiDAR more efficiently and reconstruct multi-sensory scenes with higher fidelity and scalability across diverse environments.

\begin{figure*}
  \centering
  \includegraphics[width=1.0\textwidth]{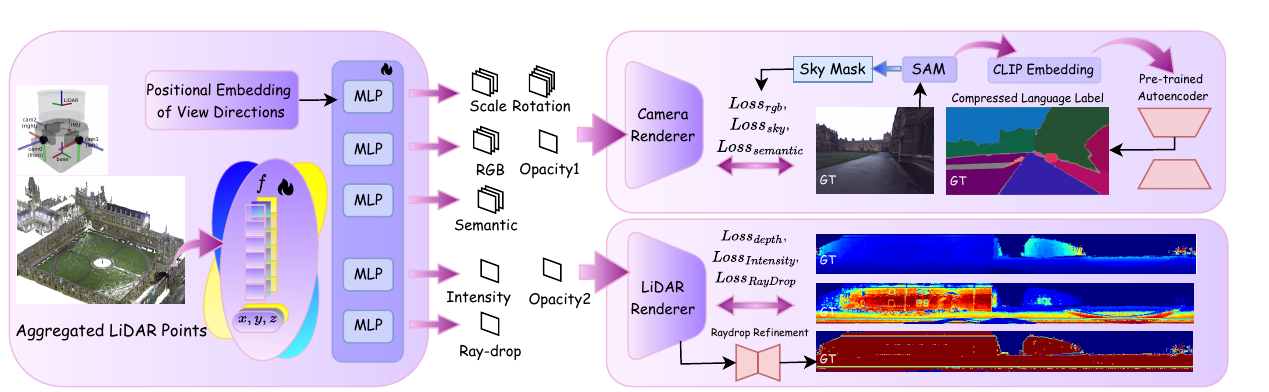}
  \caption{Pipeline of Neural-MMGS. On the left, the aggregated point clouds from the recorded LiDAR scans are used to initialise the \gls{3dgs} Gaussian set with a position and a multidimensional random embedding. Specialised decoders can decode the embedding into the parameters needed to render the three sensing modalities under consideration -- RGB, LiDAR and semantic features. These correspond not only to the sensor outputs but also to each Gaussian point's opacity, scale, and orientation. On the right, the ground-truth signals are preprocessed to provide learning supervision for the embeddings and decoder networks. In particular, the semantic masks are recovered using an autoencoder to compress the 512-dimensional CLIP \cite{radford2021clip} feature maps into 3-dimensional vectors, seeded using SAM masks \cite{kirillov2023segment}.}
  \label{fig:pipeline}
\end{figure*}

\section{Methods}

Let $\mathcal{G} = { G_i }$ denote the set of $N$ Gaussians. In traditional multimodal 3D reconstruction, each 3D Gaussian $G_i$ is typically associated with an extensive set of modality-specific parameters, leading to a large memory footprint. These parameters include the spatial position $s_i \in \mathbb{R}^3$ and covariance $\Sigma_i \in \mathbb{R}^{3 \times 3}$ (symmetric, stored as 6 values), the RGB appearance components such as color $c_i \in \mathbb{R}^3$, opacity $\alpha_i \in \mathbb{R}$, and a set of $h$ spherical harmonics for modeling view-dependent shading (commonly $3h$ for RGB). For LiDAR-specific modelling, additional parameters include LiDAR opacity, intensity $i_i \in \mathbb{R}$, spherical harmonics for intensity ($h$ values), and a ray-drop probability. Finally, semantic understanding requires a semantic feature vector $\sigma_i \in \mathbb{R}^d$, where $d$ denotes the embedding dimension.

Our Neural-MMGS framework dramatically reduces the parameter count by replacing all modality-specific attributes—except for the Gaussian position—with a shared, compact embedding. Specifically, each Gaussian $G_i$ only retains its position $s_i \in \mathbb{R}^3$ and a 32-dimensional general-purpose feature embedding $\mathcal{E}_i \in \mathbb{R}^{32}$. This embedding encapsulates appearance, geometric, and semantic information in a unified form.
Modality-specific decoders are then trained to decode $\mathcal{E}_i$ into the necessary parameters for rendering across RGB, LiDAR, and semantic modalities.

For example, with $h = 9$ (3rd-order SH) and $d = 16$, a traditional Gaussian stores approximately $68$ parameters, while Neural-MMGS's Gaussians require only 35 parameters.

These parameters and decoders are jointly learned from multimodal sensor inputs (RGB, LiDAR, semantics) and their poses. Each observation contributes position $\mathbf{P} \in \mathbb{R}^3$ and direction $\mathbf{v} \in \mathbb{R}^3$, enabling novel view synthesis across modalities. This design allows Neural-MMGS to achieve accurate and scalable multimodal rendering. An overview of our pipeline is shown in \cref{fig:pipeline}.

\subsection{LiDAR and Semantic Maps Definitions}
While RGB images can be used directly for the training and rendering processes, we need to adapt sparse LiDAR scans to a dense reconstruction task and extract meaningful semantic maps.

\paragraph{LiDAR Image Construction}

Given a LiDAR scan at time $t$, $\mathcal{P}_t = \{ (p_i^t, i_i^t) | p_i^t \in \mathbb{R}^3, i_i^t \in \mathbb{R} \}$, where $p_i^t$ is the position in Cartesian space of the $i$-point and $i_i^t$ its associated intensity.
To convert this sparse 3D data into a dense 2D structure, we project each point onto a range image using its spherical coordinates.

Specifically, we compute the azimuth \(\phi\) and elevation \(\theta\) of each point and assuming a desired image resolution of \(W \times H\), where \(H\) is determined by the number of vertical laser beam channels and \(W\) is a predefined horizontal resolution, we map each point to a pixel coordinate \((u, v)\) via:
\[
u = \left\lfloor \frac{W}{2\pi} (\phi + \pi) \right\rfloor, \quad
v = \left\lfloor \frac{H}{\Theta_{\max} - \Theta_{\min}} (\Theta_{\max} - \theta) \right\rfloor,
\]
where \(\Theta_{\min}\) and \(\Theta_{\max}\) denote the LiDAR's minimum and maximum vertical field-of-view angles.

In this way, we extract three different LiDAR images: intensity, range and raydrop images.
For each pixel, they contain, respectively, the raw intensity values, the distance value, and a binary mask representing valid or invalid measurements, i.e., the probability that the LiDAR beam is occluded or absorbed before return.

% \[
% \phi = \arctan2(y, x), \quad \theta = \arcsin\left(\frac{z}{d}\right), \quad d = \sqrt{x^2 + y^2 + z^2}.
% \]
%

\paragraph{Hierarchical Low-Dimensional Semantic Label Generation}
For the generation of semantic maps, we follow LangSplat \cite{qin2024langsplat}: we extract image segments using a pre-trained SAM \cite{kirillov2023segment} model and describe each pixel with a 512-dimensional language embedding from a pre-trained CLIP model \cite{radford2021clip}.
We pre-train a lightweight autoencoder to compress these high-dimensional language features into scene-specific, low-dimensional language embeddings, which serve as ground-truth semantic labels for our Gaussian training process. 
The resulting semantic features can be grouped into three hierarchical levels, denoted as \textit{small}, \textit{medium} and \textit{large}.
In particular, following \cite{qin2024langsplat}, the dimension of the compressed representations is $\mathbb{R}^3$. 

\subsection{General Neural Gaussian Representation}

% The reconstructed scene in represented by a set of point cloud.
Given a \gls{3dgs} scene comprising $\mathcal{G}_i$ Gaussians with their embeddings, we can render the scene from pose and direction $\mathbf{P}$ and $\mathbf{v}$ in each sensor modality through specialised, light-weight \glspl{mlp}.
The decoders retrieve not only the colour, intensity and semantic embedding, but also intermediate properties like covariance and opacity.
% For each initialised LiDAR point, we pair it with a unified embedding $ \mathcal{E}$, which includes its optical, semantic and some physical properties. 
% The embedding is feeded forward into several light weighted Multi-Layer Perceptrons (MLPs) to be decoded into specific renderable camera, lidar, and semantic parameters. 

\paragraph{3D Covariance Decoder}
Given a Gaussian point $G_i$ with its embedding $\mathcal{E}_i$ and position $s_i$, we decode the covariance using an \gls{mlp} $\mathrm{MLP}_\Sigma$ as in:
\begin{equation}
% \mathrm{\Sigma}_i = \mathrm{MLP}_{\Sigma}(\mathcal{E}_i, s_i), 
\mathrm{\Sigma}_i = \mathrm{MLP}_{\Sigma}(\mathcal{E}_i).
\end{equation}
the covariance matrix $\Sigma = R(\theta)^T \mathrm{diag(s)}R(\theta)$ is composed of a scaling matrix $s$ and a rotation matrix $R(\theta)$ represented via quaternions

\paragraph{Opacity Decoders}
We predict two separate opacity values to account for the fundamentally different penetration behaviours of light and LiDAR signals --  \(\alpha_i \) and \(\beta_i\), respectively.
These are decoded through the \glspl{mlp} $\mathrm{MLP}_\alpha$ and $\mathrm{MLP}_\beta$ based solely on spatial location, as in: 
The first opacity \(\alpha_i \), used for colour rendering, reflects the radiance field opacity under visible light, and is predicted based solely on spatial location:
\begin{equation}
\alpha_i = \mathrm{MLP}_{\alpha}(\mathcal{E}_i, s_i), \qquad \beta_i = \mathrm{MLP}_{\beta}(\mathcal{E}_i, s_i)
\end{equation}

\paragraph{Positional Embedding for View Direction}
Each Gaussian ellipsoid's scale, shape, colour, and intensity vary with the viewing direction \(\mathbf{v} \) due to anisotropic appearance and shading effects.
To model such view-dependent characteristics, traditional \glspl{3dgs} store Spherical Harmonics within each Gaussian to extract from the low-dimensional and smooth $\mathbf{v}$ a higher-dimensional representation.
Whilst this representation can express high-frequency variations in appearance, particularly in the presence of complex specular or reflective surfaces, it also challenges memory usage.

For this reason, we adopt a $L$-ranked \gls{pe} to achieve the same scope at a lesser memory burden.
We define the \gls{pe} as:
\begin{equation}
\begin{aligned}
\mathrm{PE}(\mathbf{v}) = {} & [\sin(2^0\pi \mathbf{v}), \cos(2^0\pi \mathbf{v}), \ldots, \\
                             & \sin(2^{L-1}\pi \mathbf{v}), \cos(2^{L-1}\pi \mathbf{v})]
\end{aligned}
\end{equation}
and use it as input to the decoders.
For simplicity in the notation, from now on, when we mention $\mathbf{v}$ as the input to a decoder, we refer to $\mathrm{PE}(\mathbf{v})$.

In our setting, the covariance, opacity, opacity for intensity, and semantic can be decoded from only the position and the unified embedding of the point. But the RGB colour, intensity, and raydrop also relate to the rendering viewpoint. 

\paragraph{Color Decoder}
Given $\mathcal{E}_i$, $s_i$ and the view direction $\mathbf{v}$, we can decode the Colour in a view-dependent manner using an \gls{mlp} $\mathrm{MLP}_c$, as in:
\begin{equation}
c_i = \mathrm{MLP}_{c}(\mathcal{E}_i, \mathbf{v}, \| s_i - \mathbf{P} \|)
\end{equation}
where \(\| s_i - \mathbf{P} \|\) is the Euler distance of the current Gaussian $G_i$ from the viewing position.

\paragraph{Intensity and Ray-Drop Decoders}

Inspired by the physical modelling of LiDAR signal return, we design our MLP-based decoders with physically meaningful inputs.
Specifically, the LiDAR return intensity \( i_i \) is known to be directly influenced by surface reflectivity and incident angle, and inversely proportional to distance from the sensor.
% A commonly used physical approximation is:
% \[
% i_i \propto \frac{\rho \cdot \cos(\theta)}{d^2},
% \]
% where \( \rho \) is the surface reflectivity, \( \theta \) is the angle between the incoming laser beam and the surface normal, and \( d \) denotes the Euclidean distance from the observed point to the LiDAR center.
% 
Motivated by this, we predict intensity and ray-drop probability using the \glspl{mlp} $\mathrm{MLP}_{i}$ and $\mathrm{MLP}_r$ as in:

\begin{align}
i_i &= \mathrm{MLP}_{i}(\mathcal{E}_i, \mathbf{v}, \| s_i - \mathbf{P} \|, \mathbf{e}), \\
r_i &= \mathrm{MLP}_{r}(\mathcal{E}_i, \mathbf{v}, \| s_i - \mathbf{P} \|, \mathbf{e})
\end{align}

where \( \mathbf{e} \in \mathbb{R}^4 \) is an environment descriptor learned and shared for the whole scene that should capture environmental effects such as humidity and weather.
$r_i \in \mathbb{R}$, instead, is the raydrop probability and, as such, is defined in the range $[0, 1]$. 

\paragraph{Semantic Decoder}
Finally, we predict the low-dimensional semantic feature vector $\sigma_i$ from the embedding $\mathcal{E}_i$ through the \gls{mlp} $\mathrm{MLP}_\sigma$ as in:
\begin{equation}
\sigma_i = \mathrm{MLP}_{\sigma}(\mathcal{E}_i).
\end{equation}
Unlike radiance or intensity, semantic meaning is assumed to be view-invariant — a Gaussian point should represent the same object class regardless of the viewing direction.

\subsection{Semantic Rendering}
To incorporate the semantic signals during rendering, we compute the aggregated language embedding at pixel \( v \) for each level -- small, medium or large -- using $\alpha$-blending with the RGB opacity $\alpha$, as in:

\begin{equation}
\sigma_v = \sum_{i \in \Gamma} \alpha_i \sigma_i \prod_{j \in \Gamma, j \neq i} (1 - \alpha_j)
\label{eq:semantic_rendering}
\end{equation}
where $\Gamma$ denotes the set of Gaussians in $\mathcal{G}$ contributing to pixel \( v \) regulated by the covariance $\Sigma_i$.

\subsection{LiDAR Rendering}

To render Gaussians in the LiDAR range view, we also perform $\alpha$-blending to accumulate intensity \( i \) and ray-drop probability \( r \), treated as a two-channel feature, here using $\beta$.
However, we cannot assume that all the Gaussians can be splatted on the same plane, as now the LiDAR sensor is inherently polar.
For this reason, given a viewing direction, we project the covariance $\sigma_i$ for each Gaussian on its orthogonal plane.
This captures the anisotropic extent of each Gaussian and determines its influence region on the LiDAR image.

% the decoded covariance $\Sigma_i$ and extracting its components orthogonal to the viewing direction.
% Starting from the output of the covariance decoder \( \mathrm{MLP}_{\text{cov}} \), we obtain the predicted 3D axis-aligned scales \( \boldsymbol{s} \in \mathbb{R}^3 \).
% 
% Given a unit viewing direction \( \mathbf{v} \in \mathbb{R}^3 \), we compute the projection matrix \( \mathbf{P} = \mathbf{I}_3 - \mathbf{v}\mathbf{v}^\top \), which projects vectors onto the plane orthogonal to \( \mathbf{v} \). The projected 2D covariance is then computed as:
% \[
% \boldsymbol{\Sigma}_{2D} = \mathbf{P} \boldsymbol{\Sigma}_{3D} \mathbf{P}^\top.
% \]
% We extract the two largest eigenvalues of \( \boldsymbol{\Sigma}_{2D} \) and take their square roots to represent the 2D scale of the Gaussian in the view-aligned plane.
% This projection allows us to capture the anisotropic spatial extent of each Gaussian on the LiDAR range image and provide us with each Gaussian's range of affected pixels.
% We can then apply $\alpha$-blending to accumulate intensities and raydrop probabilities, which are combined as a two-channel feature, suing the opacity $\beta$.

\subsection{Loss Function}
\paragraph{RGB Loss} % To supervise the eRGB rendering of the \gls{3dgs}
We use SAM to generate a sky mask \( M_{\text{sky}} \) for each image, where sky pixels are 1 and non-sky pixels are 0. To avoid reconstructing the sky region, we compute the RGB loss only on non-sky pixels:

\begin{align}
L_{\text{RGB}} = & (1 - \lambda_{\text{DSSIM}}) \cdot \| (1 - M_{\text{sky}}) \odot (I - I_{\text{gt}}) \|_1 \notag \\
& + \lambda_{\text{DSSIM}} \cdot \| (1 - M_{\text{sky}}) \odot (1 - \text{SSIM}(I, I_{\text{gt}})) \|_1
\end{align}

\begin{figure}[t]
    \centering

    % Left column: Observatory Quarter
    \begin{minipage}[t]{0.48\textwidth}
        \centering
        \includegraphics[width=\textwidth]{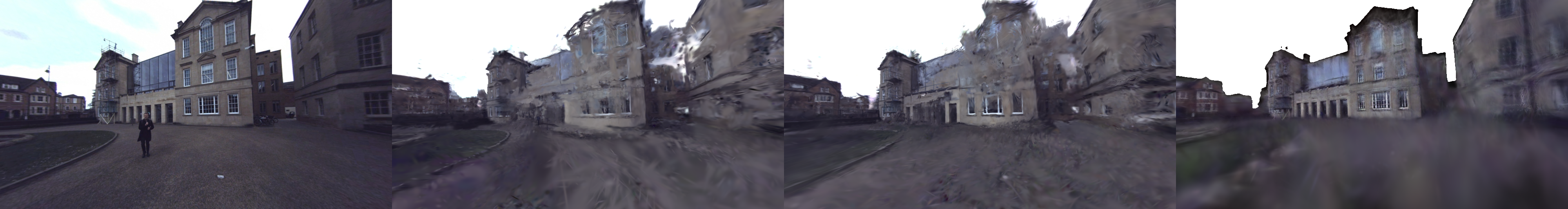} \\[0.3em]
        \includegraphics[width=\textwidth]{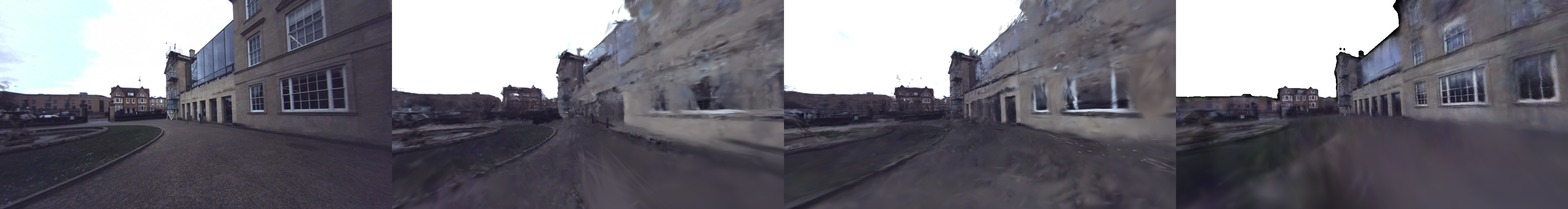} \\[0.3em]
        \includegraphics[width=\textwidth]{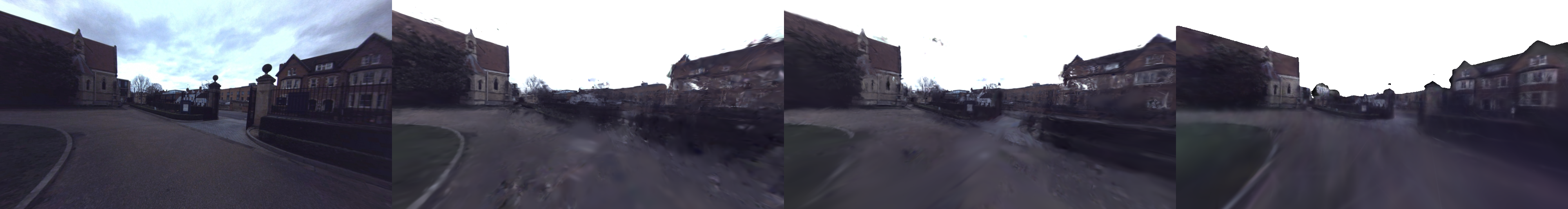}
    \end{minipage}
    \hfill
    % Right column: Keble College
    \begin{minipage}[t]{0.48\textwidth}
        \centering
        \includegraphics[width=\textwidth]{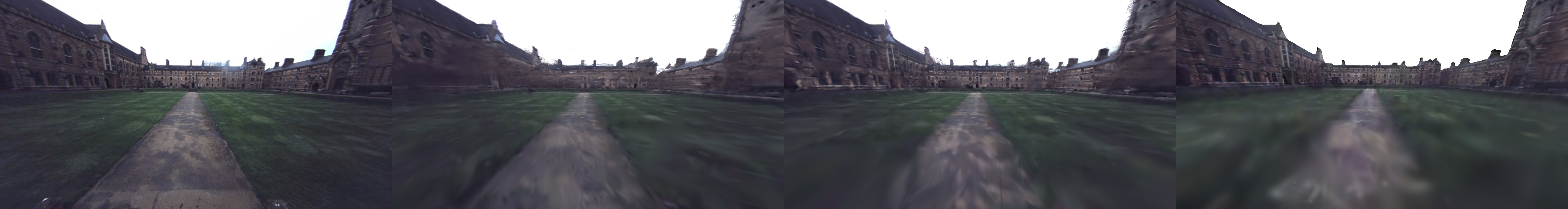} \\[0.3em]
        \includegraphics[width=\textwidth]{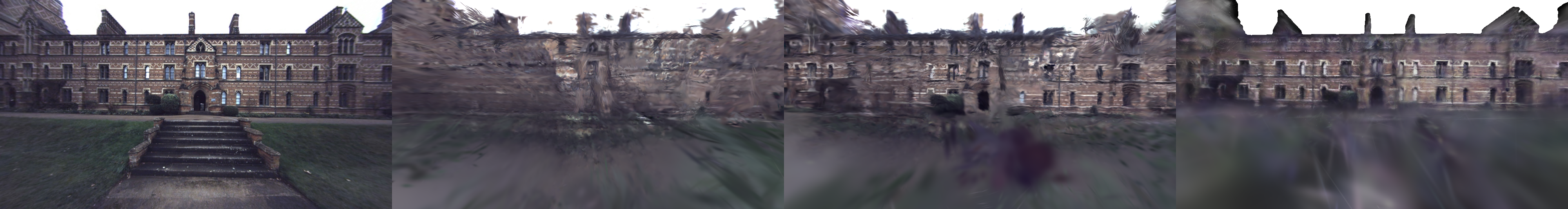} \\[0.3em]
        \includegraphics[width=\textwidth]{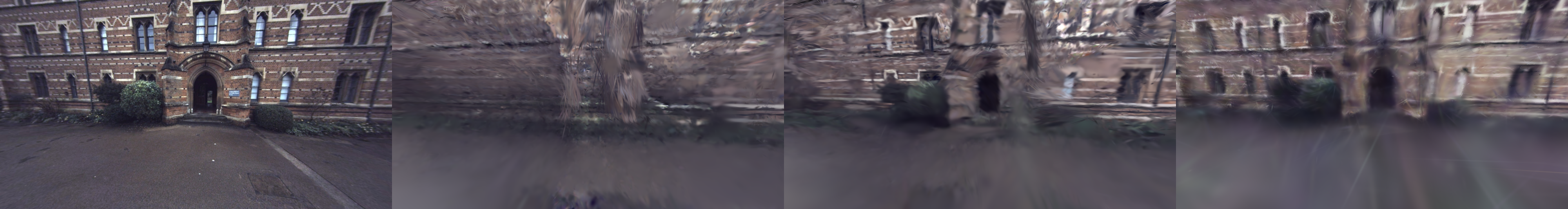}
    \end{minipage}

    \caption{Qualitative comparison on two large-scale Oxford Spires out-sequence datasets.  
    \textbf{Left column:} Reconstruction results on the \textit{Observatory Quarter} sequence using Ground Truth, Splatfacto, Splatfacto-big, and Neural-MMGS.  
    \textbf{Right column:} Results on the \textit{Keble College} sequence using Ground Truth, Splatfacto, Splatfacto-big, and Neural-MMGS.}
    \label{fig:combined_comparison}
\end{figure}

where \(I\) and \(I_{\text{gt}}\) denote the predicted and ground truth images respectively, and \(\odot\) denotes element-wise multiplication. Note that \(\text{SSIM}(I, I_{\text{gt}})\) refers to the per-pixel SSIM map, and the mask \(1 - M_{\text{sky}}\) excludes sky pixels from the loss calculation.

\begin{table*}[htbp]
\centering
\caption{\textbf{Quantitative evaluation of Novel View Synthesis on the Spires.}}

\resizebox{0.8\textwidth}{!}{\begin{tabular}{ll|ccc|ccc}
\toprule
\textbf{Sequence} & \textbf{Method} & \multicolumn{3}{c|}{\textbf{In-Sequence}} & \multicolumn{3}{c}{\textbf{Out-of-Sequence}} \\
 & & PSNR $\uparrow$ & SSIM$\uparrow$ & LPIPS$\downarrow$ & PSNR$\uparrow$ & SSIM$\uparrow$ & LPIPS$\downarrow$ \\
\midrule

\multirow{4}{*}{Observatory Quarter} 
& Nerfacto       & 23.40 & 0.807 & 0.336 & 21.25 & 0.786 & 0.370 \\
& Nerfacto-big   & 20.66 & 0.807 & \textbf{0.292} & 19.38 & \textbf{0.787} & \textbf{0.317} \\
& Splatfacto     & 22.76 & 0.791 & 0.373 & 19.47 & 0.736 & 0.445 \\
& Splatfacto-big & \textbf{23.54} & \textbf{0.811} & 0.347 & 20.26 & 0.761 & 0.413 \\
& Neural-MMGS & 23.23 & 0.779 & 0.414 & \textbf{21.52} & 0.740 & 0.420\\
\midrule

\multirow{4}{*}{\shortstack[l]{Blenheim\\Palace}} 
& Nerfacto       & 18.42 & 0.716 & 0.506 & 17.09 & 0.682 & 0.537 \\
& Nerfacto-big   & 17.93 & 0.724 & \textbf{0.445} & 17.09 & 0.695 & \textbf{0.493} \\
& Splatfacto     & 19.34 & 0.726 & 0.589 & 16.02 & 0.668 & 0.659 \\
& Splatfacto-big & 19.77 & 0.733 & 0.576 & 16.20 & 0.671 & 0.643 \\
& Neural-MMGS & \textbf{21.74} & \textbf{0.743} & 0.550 & \textbf{20.51} & \textbf{0.712} & 0.560\\
\midrule

\multirow{4}{*}{Keble College} 
& Nerfacto       & 21.10 & 0.731 & 0.397 & 20.29 & \textbf{0.748} & \textbf{0.368} \\
& Nerfacto-big   & 19.71 & \textbf{0.749} & \textbf{0.326} & 18.15 & 0.736 & 0.381 \\
& Splatfacto     & 20.47 & 0.651 & 0.514 & 19.92 & 0.658 & 0.500 \\
& Splatfacto-big & 21.36 & 0.688 & 0.478 & 20.86 & 0.707 & 0.434 \\
& Neural-MMGS & \textbf{22.34} & 0.691 & 0.510  & \textbf{21.325} & 0.687 & 0.466 \\
\bottomrule
\label{tab:NVS on Spires}
\end{tabular}}
\end{table*}

\paragraph{Semantic Loss}
We extract language feature maps and compute the semantic loss only on the valid regions (excluding sky):

\begin{equation}
L_{\text{semantic}} = \lambda_{\text{semantic}} \cdot | (1 - M_{\text{sky}}) \odot (S - S_{\text{gt}}) |_1,
\end{equation}
where $S$ and $S_{\text{gt}}$ are the predicted and ground truth three dimentional language features, masked by the valid region mask.

\paragraph{LiDAR Loss}  
For LiDAR supervision, we include three terms: \textit{depth loss}, \textit{intensity loss}, and \textit{raydrop loss}. All losses are computed only over valid LiDAR ray regions, as indicated by the binary ray drop mask \( M_{\text{ray}} \), where a value of 1 denotes a valid ray.

The depth loss compares the predicted range-view depth \( D \) with the ground truth depth \( D_{\text{gt}} \), masked by \( r_{\text{ray\_gt}} \):
\begin{equation}
L_{\text{depth}} = \| r_{\text{ray\_gt}} \odot (D - D_{\text{gt}}) \|_1.
\end{equation}

The intensity loss is a weighted combination of L1 loss and structural dissimilarity (1 - SSIM), also computed only on valid rays:
\begin{multline}
L_{\text{intensity}} = \lambda_{\text{intensity}} \cdot \Big[
(1 - \lambda_{\text{DSSIM}}^{\text{int}}) \cdot 
\| r_{\text{ray\_gt}} \odot (i_{\text{lidar}} - i_{\text{lidar\_gt}}) \|_1 \\
+ \lambda_{\text{DSSIM}}^{\text{int}} \cdot 
\| r_{\text{ray\_gt}} \odot (1 - \text{SSIM}(i_{\text{lidar}}, i_{\text{lidar\_gt}})) \|_1
\Big]
\end{multline}

The raydrop loss penalizes discrepancies between the predicted and ground truth raydrop probabilities using mean squared error:
\begin{equation}
L_{\text{raydrop}} = \text{MSE}(r_{\text{ray}}, r_{\text{ray\_gt}}).
\end{equation}

Here, \( \odot \) denotes element-wise multiplication. All supervision is restricted to regions where the LiDAR signal is valid, ensuring that the model is not penalized for uncertain or missing data.

\begin{table*}[htbp]
\centering
\caption{\textbf{LiDAR simulation comparison on KITTI-360 \textit{Static} Scene Sequence.}}
\label{lidar simu}
\small
\resizebox{\textwidth}{!}{
\begin{tabular}{l|ccccc|ccccc}
\toprule
\textbf{Method}  & \multicolumn{5}{c|}{\textbf{Depth}} & \multicolumn{5}{c}{\textbf{Intensity}} \\
& RMSE$\downarrow$ & MedAE$\downarrow$ & LPIPS$\downarrow$ & SSIM$\uparrow$ & PSNR$\uparrow$ 
& RMSE$\downarrow$ & MedAE$\downarrow$ & LPIPS$\downarrow$ & SSIM$\uparrow$ & PSNR$\uparrow$ \\
\midrule
LiDARsim ~\cite{manivasagam2020lidarsim} & 6.5470 & 0.0759 & 0.2289 & 0.7157 & 21.7746 & 0.1532 & 0.0506 & 0.2502 & 0.4479 & 16.3045 \\
NKSR ~\cite{huang2023NKSR} & 4.6647 & 0.0698 & 0.2295 & 0.7052 & 22.5390 & 0.1565 & 0.0536 & 0.2429 & 0.4200 & 16.1159 \\
PCGen ~\cite{li2023pcgen} & 4.8838 & 0.1785 & 0.5210 & 0.5062 & 24.3050 & 0.2005 & 0.0818 & 0.6100 & 0.1248 & 13.9606 \\
LiDAR-NeRF ~\cite{tao2024lidarnerf}& 3.6801 & 0.0667 & 0.3523 & 0.6043 & 26.7663 & 0.1557 & 0.0549 & 0.4212 & 0.2768 & 16.1683 \\
LiDAR4D~\cite{zheng2024lidar4d} & 3.2370 & 0.0507 & 0.1313 & 0.7218 & 27.8840 & 0.1343 & 0.0404 & 0.2127 & 0.4698 & 17.4529 \\
GS-LiDAR ~\cite{jiang2025gslidar} & \textbf{2.8895} & \textbf{0.0411} & \textbf{0.0997} & \textbf{0.8454} & 28.8807 & \textbf{0.1211} & \textbf{0.0359} & \textbf{0.1630} & \textbf{0.5756} & \textbf{18.3506} \\
Neural-MMGS  & 2.9096 & 0.0894 & 0.1458 & 0.8055 & \textbf{28.9101} & 0.1418 & 0.0606 & 0.3316 & 0.4277 & 16.9736 \\
\bottomrule
\end{tabular}
}
\end{table*}

\paragraph{Sky Loss} Although sky pixels are masked out during training, some Gaussian ellipsoids may still partially splat into sky regions. To mitigate this issue, we introduce a sky loss that penalizes the accumulated opacity of Gaussians rendered onto sky pixels, encouraging it to remain close to zero:
\begin{equation}
L_{\text{sky}} = \lambda_{\text{sky}} \cdot \|\alpha_{\text{sky}}\|_2^2,
\end{equation}
where $\alpha_{\text{sky}}$ denotes the accumulated alpha values in the sky region.

We combine the depth, intensity, and raydrop losses into a single LiDAR supervision term:
\begin{equation}
L_{\text{lidar}} = L_{\text{depth}} + L_{\text{intensity}} + L_{\text{raydrop}}.
\end{equation}

Finally, the overall training loss is defined as the weighted sum of all individual components:
\begin{align}
L_{\text{total}} = & \ \lambda_{\text{RGB}} \cdot L_{\text{RGB}} 
+ \lambda_{\text{semantic}} \cdot L_{\text{semantic}} \notag \\
& + \lambda_{\text{lidar}} \cdot L_{\text{lidar}} 
+ \lambda_{\text{sky}} \cdot L_{\text{sky}}
\end{align}
where each $\lambda$ term controls the relative importance of the corresponding loss component.

\section{Experiments}

%   \begin{figure*}[htbp]
%     \centering
%     \includegraphics[width=0.8\textwidth]{Figures/Kitti360/LiDAR_Depth_Comparison.pdf} 
%     \caption{LiDAR Depth Reconstruction Results Comparison with Baselines.}
%     \label{fig:lidar_depth_comparison}
% \end{figure*}

%   \begin{figure*}[htbp]
%     \centering
%     \includegraphics[width=0.8\textwidth]{Figures/Kitti360/LIiDAR_Intensity_Comparison.pdf} 
%     \caption{LiDAR Intensity Reconstruction Results Comparison with Baselines.}
%     \label{fig:lidar_intensity_comparison}
% \end{figure*}

  \begin{figure*}[htbp]
    \centering
    \includegraphics[width=\textwidth]{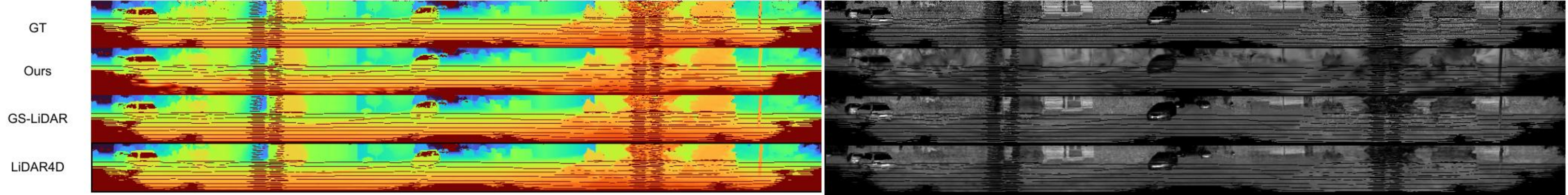} 
    \caption{LiDAR Depth and Intensity Reconstruction Results Comparison with Baselines.}
    \label{fig:lidar_intensity_comparison}
\end{figure*}

\paragraph{Datasets}
We test our method on Oxford Spires Dataset ~\cite{tao2024oxford} and KITTI-360 ~\cite{liao2022kitti}, both of which employ 64-beam LiDAR sensors and surround-view cameras.
The Spires Dataset captures scenes of well-know landmarks in Oxford, suitable for novel-view synthesis.
We use the following three sequences: Observatory Quarter, Blenheim Palace, and Keble College.
We split the dataset into in-sequence set and out-sequence set following the Spires novel-view synthesis benchmark~\cite{tao2024oxford}.
The out-sequence views are sampled from trajectories that are quite different from the training poses, which are much more challenging.
KITTI-360, on the other hand, is an autonomous driving dataset, which includes traversals of urban areas in Karlsruhe, Germany.
Following previous methods~\cite{zheng2024lidar4d, jiang2025gslidar}, we select static sequences for LiDAR NVS evaluation.

\paragraph{Evaluation Metrics}
To evaluate the multi-modal reconstruction capability of Neural-MMGS, we assessed its performance on three key tasks: RGB image novel-view synthesis, LiDAR depth reconstruction, and LiDAR intensity reconstruction. For image novel-view synthesis, we employed Peak Signal-to-Noise Ratio (PSNR), Structural Similarity Index Measure (SSIM), and Learned Perceptual Image Patch Similarity (LPIPS) as evaluation metrics. In the case of LiDAR depth and intensity reconstruction, we utilized Root Mean Square Error (RMSE), Median Absolute Error (MedAE), LPIPS, SSIM, and PSNR to comprehensively quantify reconstruction accuracy and perceptual quality.

\paragraph{Settings}

All the experiments are done using a single RTX 4090 GPU. The reported experiment results on all the tasks are trained on our model for 7000 iterations.

\subsection{Camera Novel-view Synthesis}

% \begin{figure}[htbp]
%     \centering
%     \includegraphics[width=0.8\textwidth]{Figures/Keble_04_out_cam0/0003.png} \\[0.5em]
%     \includegraphics[width=0.8\textwidth]{Figures/Keble_04_out_cam0/0047.png} \\[0.5em]
%     \includegraphics[width=0.8\textwidth]{Figures/Keble_04_out_cam0/0053.png}
%     \caption{Reconstruction results on the Keble out-sequence dataset using different methods. From left to right: Ground Truth, Splatfacto, Splatfacto-big, and Neural-MMGS.}
%     \label{fig:keble04_comparison}
% \end{figure}

\paragraph{Baseline}
To evaluate the effectiveness of our method on novel view synthesis tasks, we compare it against several baselines on the Spires Dataset, including Nerfacto~\cite{tancik2023nerfstudio} and Splatfacto~\cite{ye2025gsplat}, as well as their large variants, Nerfacto-big and Splatfacto-big.

% \begin{figure}[htbp]
%     \centering
%     \includegraphics[width=0.8\textwidth]{Figures/BP_01_out_cam0_selected/0006.png} \\[0.5em]
%     \includegraphics[width=0.8\textwidth]{Figures/BP_01_out_cam0_selected/0016.png} \\[0.5em]
%     \includegraphics[width=0.8\textwidth]{Figures/BP_01_out_cam0_selected/0034.png}
%     \caption{Reconstruction results on the BP out-sequence dataset using different methods. From left to right: Ground Truth, Splatfacto, Splatfacto-big, and Neural-MMGS.}
%     \label{fig:keble04_comparison}
% \end{figure}

\paragraph{Implementation Details}
For the cloning, splitting and pruning of Gaussain elipsoids during training, we fix the MLPs and do inference to get the estimated scale and opacity under current training stage. Base on the inferenced scale, we set a threshold to decide when to clone or split a gaussian elipsoid. Following the Spires\cite{tao2024oxford}, we select two sets of evaluation frames. The in-sequence frames are sampled and held out from the training sequence, and the out-sequence frames are from another trajetory. We do a 3m voxel size downsampling on the aggregated lidar points from all the frames in the sequence as the initialization.

\paragraph{Performance Evaluation and Visualization}

Fig. ~\ref{fig:combined_comparison} shows the novel-view synthesis results on Observatory Quarter. Obviously the novel-view synthesised image generally has higher quality, especially on representing further buildings, which shows the benifit of having lidar supervision in aera that's quite far away. Tab. ~\ref{tab:NVS on Spires} shows our results compared with existing baseline on this Spires Dataset. We outperform them on PSNR on almost all of the test sequences. But our SSIM and LPIPS are not that good, partly because lidar points are extremely sparse in the closest area.

\subsection{LiDAR Simulation}

\paragraph{Baseline}

We evaluate our method alongside recent data-driven approaches LiDARsim ~\cite{wu2024dynamiclidarsim} and PCGen ~\cite{li2023pcgen}. Additionally, we compare our results with the perscene optimized reconstruction method NKSR ~\cite{huang2023NKSR}, LiDAR-NeRF ~\cite{tao2024lidarnerf} and the state-of-the-art method, LiDAR4D~\cite{zheng2024lidar4d}.

\paragraph{Implementation Details}  The rendered LiDAR raydrop maps are not discrete, so following GS-LiDAR \cite{jiang2025gslidar} and LiDAR4D \cite{zheng2024lidar4d}, we post-train another scene specific U-Net for 1000 iterations to refine the raydrop. Following settings from baselines, we randomly sample number of LiDAR points for point initialization. While the baselines sample $1 \times10^6$ LiDAR points, we sample only $2 \times10^5$ points due to GPU storage limitation.
% As noted, we select consecutive  static frames as a single scene and hold out samples at regular intervals for novel view synthesis (NVS) evaluation.
For evaluation, we select \textit{static} sequences as in ~\cite{zheng2024lidar4d, jiang2025gslidar}, which consist of 64 consecutive frames each, and hold out 4 samples at regular intervals for LiDAR NVS evaluation.
We undistort the fish-eye images following ~\cite{mei2007single}.

\paragraph{Performance Evaluation}
Tab.~\ref{lidar simu} shows the LiDAR simulations results of Neural-MMGS compared with baselines. Our method beats most of the baselines on depth metrics, and we obtain competitive results on intensity metrics. It should be mentioned that the models of baselines are doing this single task while we reconstruct both the camera image and lidar range-view image, which is a much harder task through joint training. More visualization results can be found in the appendix.

  % In this section we provide the visualization results of our methods doing LiDAR simulation, as well as the comparison with baselines.

\begin{figure*}[htbp]
    \centering
    \includegraphics[width=0.8\textwidth]{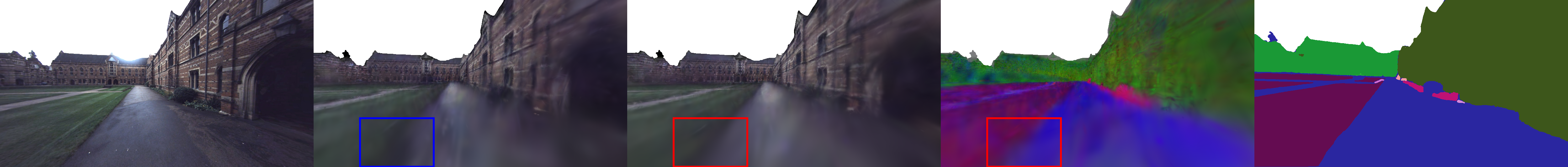} 
    \caption{RGB Reconstruction Results w/o semantic supervision. (From left to right: ground truth image, reconstructed novel-view without semantic guidance, reconstructed novel-view with semantic guidance, trained semantic label, ground truth semantic label.)}
    \label{fig:keble04_w/o_semantic}
\end{figure*}

% \begin{table*}[htbp]
% \centering
% \caption{\textbf{Ablation study on the Oxford Spires Dataset.}}
% \small
% \resizebox{0.8\textwidth}{!}{
% \begin{tabular}{l|ccc|ccc}
% \toprule
% \textbf{Method / Semantic Level} & \multicolumn{3}{c|}{\textbf{In-Sequence}} & \multicolumn{3}{c}{\textbf{Out-of-Sequence}} \\
% & PSNR $\uparrow$ & SSIM $\uparrow$ & LPIPS $\downarrow$ & PSNR $\uparrow$ & SSIM $\uparrow$ & LPIPS $\downarrow$ \\
% \midrule
% \multicolumn{7}{l}{\textit{Module Component Ablation}} \\
% Neural-MMGS                        & 22.34  & 0.691 & 0.510 & 21.33  & 0.687 & 0.466 \\
% w/o LiDAR Loss                 & 22.37  & 0.693 & 0.510 & 21.04  & 0.686 & 0.448 \\
% w/o Sky Mask                   & 21.96  & 0.690 & 0.510 & 17.68  & 0.667 & 0.505 \\
% w/o Positional Embedding       & 22.08  & 0.688 & 0.523 & 20.16  & 0.675 & 0.470 \\

% w/o Semantic Loss              & 22.089 & 0.689 & 0.512 & 21.101 & 0.682 & 0.471 \\

% \midrule
% \multicolumn{7}{l}{\textit{Semantic Supervision Level Ablation}} \\
% Small Semantic                 & 22.250 & 0.689 & 0.512 & 21.013 & 0.686 & 0.466 \\
% Medium Semantic                & 22.244 & 0.690 & 0.511 & 21.237 & \textbf{0.687} & \textbf{0.462} \\
% Large Semantic                 & \textbf{22.335} & \textbf{0.691} & \textbf{0.510} & \textbf{21.325} & \textbf{0.687} & 0.466 \\
% \bottomrule
% \end{tabular}
% }
% \label{tab:ablation-merged}
% \end{table*}

\begin{table}[H]
\centering
\caption{\textbf{Ablation study on the Oxford Spires Dataset.}}
\footnotesize
\renewcommand{\arraystretch}{1.1}
\resizebox{\columnwidth}{!}{
\begin{tabular}{l|ccc|ccc}
\toprule
\textbf{Method /} & \multicolumn{3}{c|}{\textbf{In-Sequence}} & \multicolumn{3}{c}{\textbf{Out-of-Sequence}} \\
\textbf{Semantic Level} & PSNR $\uparrow$ & SSIM $\uparrow$ & LPIPS $\downarrow$ & PSNR $\uparrow$ & SSIM $\uparrow$ & LPIPS $\downarrow$ \\
\midrule
\multicolumn{7}{l}{\textit{Module Component Ablation}} \\
Neural-MMGS                        & 22.34  & 0.691 & 0.510 & 21.33  & 0.687 & 0.466 \\
w/o LiDAR Loss                 & 22.37  & 0.693 & 0.510 & 21.04  & 0.686 & 0.448 \\
w/o Sky Mask                   & 21.96  & 0.690 & 0.510 & 17.68  & 0.667 & 0.505 \\
w/o Pos Embedding       & 22.08  & 0.688 & 0.523 & 20.16  & 0.675 & 0.470 \\
w/o Semantic Loss              & 22.089 & 0.689 & 0.512 & 21.101 & 0.682 & 0.471 \\
\midrule
\multicolumn{7}{l}{\textit{Semantic Supervision Level Ablation}} \\
Small Semantic                 & 22.250 & 0.689 & 0.512 & 21.013 & 0.686 & 0.466 \\
Medium Semantic                & 22.244 & 0.690 & 0.511 & 21.237 & \textbf{0.687} & \textbf{0.462} \\
Large Semantic                 & \textbf{22.335} & \textbf{0.691} & \textbf{0.510} & \textbf{21.325} & \textbf{0.687} & 0.466 \\
\bottomrule
\end{tabular}
}
\label{tab:ablation-merged}
\end{table}

% \begin{table}[htbp]
% \centering
% \caption{\textbf{Ablation study on the Oxford Spires Dataset.}}
% \footnotesize  % 可以改成 \small 或 \normalsize
% \setlength{\tabcolsep}{4pt}  % 默认是 6pt，可以减小或增大
% \renewcommand{\arraystretch}{1.1} % 行高稍微增加，让字体看起来更舒适
% \begin{tabular}{l|ccc|ccc}
% \toprule
% \textbf{Method / Semantic Level} & \multicolumn{3}{c|}{\textbf{In-Sequence}} & \multicolumn{3}{c}{\textbf{Out-of-Sequence}} \\
% & PSNR $\uparrow$ & SSIM $\uparrow$ & LPIPS $\downarrow$ & PSNR $\uparrow$ & SSIM $\uparrow$ & LPIPS $\downarrow$ \\
% \midrule
% \multicolumn{7}{l}{\textit{Module Component Ablation}} \\
% Neural-MMGS                        & 22.34  & 0.691 & 0.510 & 21.33  & 0.687 & 0.466 \\
% w/o LiDAR Loss                 & 22.37  & 0.693 & 0.510 & 21.04  & 0.686 & 0.448 \\
% w/o Sky Mask                   & 21.96  & 0.690 & 0.510 & 17.68  & 0.667 & 0.505 \\
% w/o Positional Embedding       & 22.08  & 0.688 & 0.523 & 20.16  & 0.675 & 0.470 \\
% w/o Semantic Loss              & 22.089 & 0.689 & 0.512 & 21.101 & 0.682 & 0.471 \\
% \midrule
% \multicolumn{7}{l}{\textit{Semantic Supervision Level Ablation}} \\
% Small Semantic                 & 22.250 & 0.689 & 0.512 & 21.013 & 0.686 & 0.466 \\
% Medium Semantic                & 22.244 & 0.690 & 0.511 & 21.237 & \textbf{0.687} & \textbf{0.462} \\
% Large Semantic                 & \textbf{22.335} & \textbf{0.691} & \textbf{0.510} & \textbf{21.325} & \textbf{0.687} & 0.466 \\
% \bottomrule
% \end{tabular}
% \label{tab:ablation-merged}
% \end{table}

\subsection{Ablation Study}

% Following the main experiments, we test our pipeline on three tasks as the ablation study, including camera novel-view synthesis, LiDAR reconstruction, and open vocabulary understanding, which shows the effectness of each module for the multi-modal ability of Neural-MMGS.

% \subsubsection{Ablation on Camera Novel-view Syhthesis}

We do ablation on camera novel-view synthesis task. Tab.~\ref{tab:ablation-merged} presents an ablation study on the effect of each loss component in Neural-MMGS for novel-view synthesis on the Oxford Spires Dataset. Removing any individual component leads to a performance drop, with more significant degradation observed in the out-of-sequence setting. Notably, excluding the positional embedding slightly increases LPIPS, indicating its role in preserving fine-grained texture details.

To further evaluate the effectiveness of semantic guidance, we compare novel-view synthesis results under different settings: without semantic loss and with semantic labels of varying granularity. As shown in Tab.~\ref{tab:ablation-merged}, incorporating semantic labels consistently improves performance across most metrics. Interestingly, coarse semantic labels ("Large Semantic") outperform finer-grained ones, likely because in large-scale scenes, detailed semantics are less critical—coarser categories are sufficient to enhance the expressiveness of the 32-dimensional Gaussian features.

To further highlight the memory efficiency of our method, we report the GPU memory usage on the Keble-04 scene sequence in comparison with Splatfacto at the beginning of the training. For fair comparison, both methods are using same number of Colmap points as GS initialisation. As shown in Tab.~\ref{storage}, our method requires only 6598.49 MB of GPU memory. This represents a $\sim$28\% reduction in memory consumption, achieved without compromising reconstruction quality.

\begin{table}[H]
    \centering
    \caption{GPU Memory Usage Comparison}
    \begin{tabular}{lc}
        \toprule
        Method & GPU Memory Usage (MB) \\
        \midrule
        Splatfacto & 9164.57 \\
        Ours       & 6598.49 \\
        \bottomrule
    \end{tabular}
    \label{storage}
\end{table}

Fig.~\ref{fig:keble04_w/o_semantic} illustrates how semantic supervision improves RGB reconstruction. The model with semantic guidance yields sharper boundaries, particularly along roads, where semantic labels provide strong structural cues.

\section{Conclusion and Limitations}
This paper presented Neural-MMGS, a multimodal neural \gls{3dgs} framework that effectively integrates RGB images, LiDAR point clouds, and semantic features through a compact per-Gaussian embedding. By fusing optical, physical, and semantic cues into a unified representation, Neural-MMGS enables efficient and scalable reconstruction of large-scale scenes. Our modality-specific decoders support accurate synthesis and reconstruction across visual, geometric, and semantic domains, while reducing memory overhead compared to existing approaches that rely on explicit parameter concatenation. 

\paragraph{Limitations and future work}
Neural-MMGS currently cannot handle dynamic scene reconstruction, which is common in real-world robotic applications.
This is a limitation of the formulation, which would need restructuring to extend to the fourth dimension~\cite{wu2024fourdgs}.
This could be done by incorporating a conditioning on time to the decoders, and allowing the position of the Gaussians to vary over time.

Moreover, during our experiments, we noticed that the semantic embeddings provided by CLIP~\cite{radford2021clip} do not perform exceptionally well in outdoor environments.
For this reason, we reserve the right to test more modern open vocabulary models and test their application in semantic understanding.

Finally, we plan to use the proposed approach in a robotic application to simulate a robot's full sensor suite and learn navigation and control strategies leveraging reinforcement learning techniques.

% \newpage

\bibliographystyle{unsrt}  % 或其他 natbib 兼容的样式，比如 abbrvnat, unsrtnat 等
\bibliography{cite} 

\newpage

\appendix

  \section{Preliminaries}
  \subsection{Preliminaries on 3D Gaussian Splatting}

  3DGS~\cite{kerbl2023gaussiansplatting} uses anisotropic Gaussians whose shape, color, opacity and refraction parameters are estimated via multi-view image-based optimization. Each Gaussian is defined by a covariance matrix $\Sigma$, a center position $\mathbf{p}$, opacity $\alpha$, and spherical harmonics coefficients for color $\mathbf{c}$. The covariance matrix $\Sigma$ is decomposed into a scaling matrix and a rotation matrix to facilitate differentiable optimization. For rendering, all Gaussians in the view frustum are splatted using onto a 2D plane followed by $\alpha$-blending of the color based on Gaussian opacity to achieve the final image. The color $\mathbf{C}$ of a pixel is computed by blending $N$ ordered 2D Gaussians that overlap the pixel, formulated as:

\begin{equation}
\mathbf{C} = \sum_{i \in N} \mathbf{c}_i \alpha_i \prod_{j=1}^{i-1}(1 - \alpha_j)
\label{eq:color_blending}
\end{equation}

Here, the final color $\mathbf{c}_i$ of a pixel is defined by the color of each Gaussian, dependent on the spherical harmonics based on the view direction, falling on to the specific pixel multiplied by its individual opacity and the cumulative opacity $\alpha_i$ of the $i$ Gaussians in front. This approach ensures that the contributions of overlapping Gaussians are blended accurately, resulting in a precise and realistic rendered image of the scene. The implementation also allows gradients to pass to an unrestricted number of Gaussians at any depth behind the camera position. Over the course of optimization, numerous tiny changes based on the gradients from differential rendering of multi-view images allow for the Gaussians to approximate shapes and colors close to real scene geometry.

  \section{Implementation Details}

  \subsection{Detail for in LiDAR Rendering}

For the input of LiDAR renderer written in cuda, we first get the decoded 3D covariance $\Sigma_i$ and extract its components orthogonal to the viewing direction.
Starting from the output of the covariance decoder \( \mathrm{MLP}_{\text{cov}} \), we obtain the predicted 3D axis-aligned scales \( \boldsymbol{s} \in \mathbb{R}^3 \).

Given a unit viewing direction \( \mathbf{v} \in \mathbb{R}^3 \), we compute the projection matrix \( \mathbf{P} = \mathbf{I}_3 - \mathbf{v}\mathbf{v}^\top \), which projects vectors onto the plane orthogonal to \( \mathbf{v} \). The projected 2D covariance is then computed as:
\[
\boldsymbol{\Sigma}_{2D} = \mathbf{P} \boldsymbol{\Sigma}_{3D} \mathbf{P}^\top.
\]
We extract the two largest eigenvalues of \( \boldsymbol{\Sigma}_{2D} \) and take their square roots to represent the 2D scale of the Gaussian in the view-aligned plane.
This projection allows us to capture the anisotropic spatial extent of each Gaussian on the LiDAR range image and provide us with each Gaussian's range of affected pixels.
We can then apply $\alpha$-blending to accumulate intensities and raydrop probabilities, which are combined as a two-channel feature, suing the opacity $\beta$.

  \subsection{Network Structures}

\begin{figure*}[htbp]
    \centering
    \includegraphics[width=0.7\textwidth]{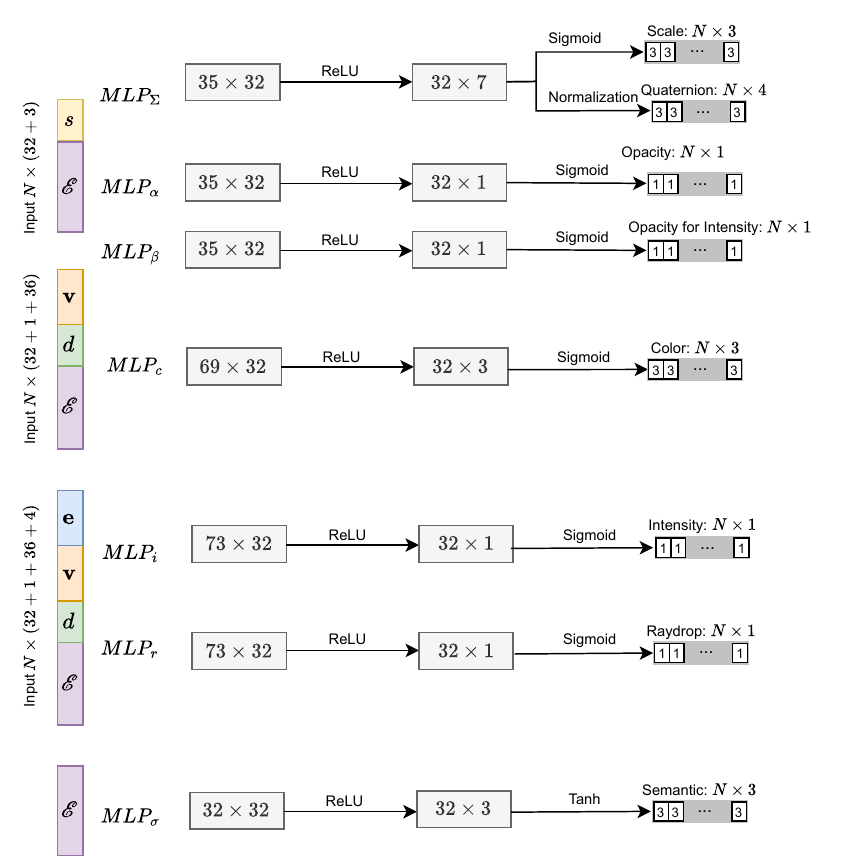} 
    \caption{\textbf{MLP Structures.} For each point, we use small
MLPs ($MLP_{\Sigma}$, $MLP_{\alpha}$, $MLP_{\beta}$, $MLP_{c}$, $MLP_i$, $MLP_r$, $MLP_{\sigma}$) to predict attributes (covariance, opacity, opacity for intensity, color, intensity, raydrop, semantic) of $k$ neural Gaussians. The input to MLPs are
feature embedding $\mathcal{E}$, point position $s$, the positional embedding of relative viewing direction $\mathbf{v}$, distance $d$ between the camera and anchor point and an environmental factor $\mathbf{e}$.}
    \label{fig:MLP_structure}
\end{figure*}

We design several MLPs as the decoders. The core MLPs include the
covariance \gls{mlp} $MLP_{\Sigma}$, the opacity \gls{mlp}s $MLP_{\alpha}$, $MLP_{\beta}$, the color \gls{mlp} $MLP_{c}$, the intensity \gls{mlp} $MLP_i$, the raydrop \gls{mlp} $MLP_r$, and the semantic \gls{mlp} $MLP_{\sigma}$. All of these \gls{mlp}s are implemented in a LINEAR $\rightarrow$ RELU $\rightarrow$ LINEAR style with the hidden dimension of 32, as illustrated in Fig.~\ref{fig:MLP_structure} Each branch’s output is activated with a head layer.

  \subsection{Other Details}

  All experiments were performed on a machine running Ubuntu 20.04 with an NVIDIA RTX 4090 GPU, using Python 3.8.10, PyTorch 1.13.1, and CUDA 11.6.
  
  \section{Additional Experiment Results}
  \subsection{Additional Experiment Results on 3D Camera Novel-view Synthesis}

  In this section, we provide more visualization results on the other sequence of Oxford Spires Dataset:the Blenheim Palace Sequence.

\begin{figure*}[htbp]
\centering
    \includegraphics[width=0.8\textwidth]{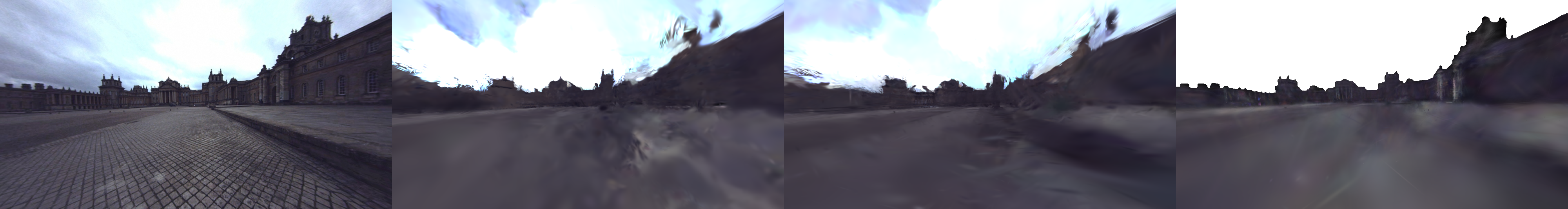} \\[0.5em]
    \includegraphics[width=0.8\textwidth]{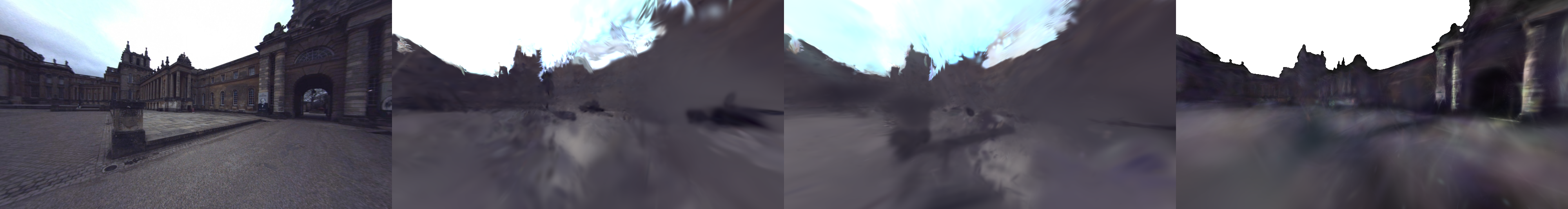} \\[0.5em]
    \includegraphics[width=0.8\textwidth]{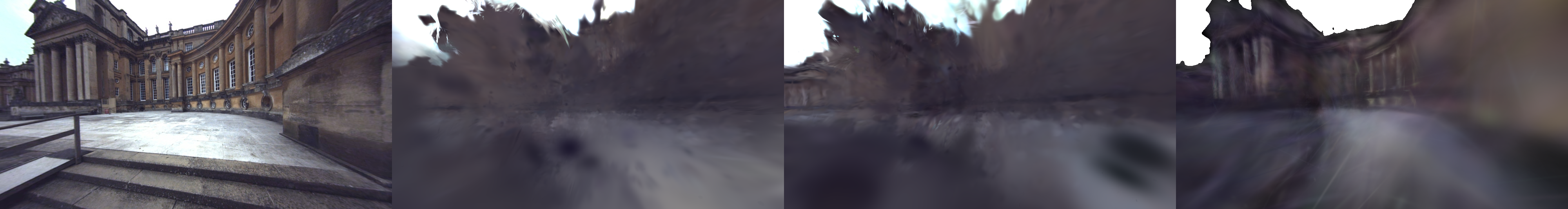}
    \caption{Reconstruction results on the Blenheim Palace out-sequence dataset using different methods. From left to right: Ground Truth, Splatfacto, Splatfacto-big, and Neural-MMGS.}
    \label{fig:BP_comparison}
\end{figure*}

\subsection{Per-point Storage}

To evaluate the storage efficiency of our neural Gaussian representation, we calculate the per-point storage of Neural-MMGS and compare it with a version of the model that explicitly represents all parameters (including camera, LiDAR, and semantic information) for each Gaussian. The per-point storage is computed by dividing the size of the checkpoint file by the number of points after 7000 training iterations. Neural-MMGS achieves a per-point storage of $1.45 \times 10^{-4}$, whereas the explicit representation requires $2.44 \times 10^{-4}$. This demonstrates that our method reduces per-point storage by 40.6\%.

  \section{Broader Impacts}

Our proposed Neural-MMGS framework enables efficient and scalable multimodal 3D scene reconstruction, with potential benefits in autonomous driving, urban planning, and AR/VR. By embedding image, LiDAR, and semantic features in a unified representation, the system supports rich scene understanding while significantly reducing memory consumption. This contributes to lower computational costs and energy usage, aligning with the goals of sustainable and resource-efficient AI.

However, such capabilities raise concerns regarding potential misuse in surveillance and possible biases in semantic data. We advocate for ethical usage of this technology, including privacy-aware data practices, energy-efficient computation, and fairness in semantic modeling.

\end{document}